\documentclass[conference]{IEEEtran}
\IEEEoverridecommandlockouts
\usepackage{cite}
\usepackage{amsmath,amssymb,amsfonts}
\usepackage{algorithmic}
\usepackage{graphicx}
\usepackage{textcomp}
\usepackage{subcaption}
\usepackage{float}
\usepackage{adjustbox}
\usepackage[hidelinks]{hyperref}
\usepackage{xcolor}
\def\BibTeX{{\rm B\kern-.05em{\sc i\kern-.025em b}\kern-.08em
    T\kern-.1667em\lower.7ex\hbox{E}\kern-.125emX}}
\begin{document}

\title{Genetic Algorithms for Evolution of QWOP Gaits}

\author{\IEEEauthorblockN{Zachary Jones}
\IEEEauthorblockA{\textit{Institute for Artificial Intelligence} \\
\textit{University of Georgia}\\
zachary.jones@uga.edu}
\and
\IEEEauthorblockN{Mohammad Al-Saad}
\IEEEauthorblockA{\textit{Department of Computer Science} \\
\textit{University of Georgia}\\
mohammad.alsaad@uga.edu}
\and
\IEEEauthorblockN{Ankush Vavishta}
\IEEEauthorblockA{\textit{Department of Computer Science} \\
\textit{University of Georgia}\\
av21224@uga.edu}
}

\maketitle



\section{Introduction}
QWOP is a browser-based, 2-dimensional flash game in which the player controls an Olympic sprinter competing in a simulated 100-meter race. The goal of the game is to advance the runner to the end of the 100-meter race as quickly as possible using the ‘Q’, ‘W’, ‘O’, and ‘P’ keys, which control the muscles in the sprinter’s legs. Despite the game’s simple controls and straightforward goal, it is renowned for its difficulty and unintuitive gameplay.

In this paper, we attempt to automatically discover effective QWOP gaits. We describe a programmatic interface developed to play the game, and we introduce several variants of a genetic algorithm tailored to solve this problem. We present experimental results on the effectiveness of various representations, initialization strategies, evolution paradigms, and parameter control mechanisms.

\section{Background}

\subsection{QWOP}

QWOP is a browser-based, 2-dimensional, free-to-play flash game developed by Bennett Foddy \cite{7}. The player controls an Olympic sprinter competing in a simulated 100-meter race, and the goal of the game is to advance the runner to the end of the race as quickly as possible without falling over. The game is won when the simulated sprinter reaches or exceeds the 100-meter finish line. The game is considered a loss if the sprinter’s upper body touches the ground at any time before the finish line.

Players control the simulated sprinter using the ‘Q’, ‘W’, ‘O’, and ‘P’ keys. Pressing and holding a particular key causes the muscles one part of the sprinter’s legs to contract. The ‘Q’ and ‘W’ keys control the contraction of the the sprinter’s left and right thigh muscles, while the ‘O’ and ‘P’ keys control the contraction of the left and right hamstring and calf muscles. The game allows multiple keys to be pressed at the same time. Precise timing and careful coordination of key presses are required to successfully advance the runner without falling - a task that has proven to be difficult and unintuitive for human players. Shortly after its development, QWOP achieved widespread infamy for its intentionally frustrating control system and unpredictable ragdoll physics engine \cite{2} \cite{17}.

Achieving a realistic bipedal gait is often quickly abandoned by new players. Most human players tend to resort to finding a repeatable pattern of inputs sufficient enough to scoot the sprinter to the finish line by balancing on one knee and kicking the free foot forward \cite{19}. Nevertheless, some determined players have achieved incredible proficiency at the game and found fast, stable bipedal gaits. The current world record is held by Roshan Ramachandra, who completed the game in 67 seconds \cite{18}.

\subsection{Genetic Algorithms}
\label{sec:2.2}

Genetic Algorithms are a family of search algorithms loosely inspired by the process of biological evolution. A genetic algorithm (GA) maintains an artificial \textit{population} of encoded candidate solutions. Each individual in the population is evaluated by a fitness function. The encoded solution is referred to as the individual’s \textit{genotype}, and the decoded solution is referred to as the individual’s \textit{phenotype}.

A genetic algorithm begins by initializing a population with N individuals having randomly generated genotypes. The algorithm then proceeds through successive \textit{generations}. Each generation begins by selecting two or more \textit{parents} from the population. Parents with better fitness are more likely to be selected. After parents are selected, a \textit{crossover} operation is applied to pairs of parents with probability $P_c$. The crossover operation generates $n$ new candidate solutions referred to as \textit{children}. Then a mutation operator is applied to each child with some probability $P_m$. The generation ends by selecting selecting $n$ members of the population for replacement with the new children. The GA iterates the genetic operations through generations until it reaches some terminating condition, such as a sufficiently optimized solution or a fixed number of generations.

GAs have been shown to be robust problem solvers, particularly for optimization problems where the fitness landscape might have several local optima. In this paper, we use three varieties of genetic algorithm - a generational model, a cellular model, and a steady-state model.

In a \textit{generational} GA, the entire population is replaced by a new population each iteration. In this type of GA, no replacement mechanism need be specified, since all children survive to replace all parents. Compared to the other varieties, generational GAs may converge more slowly, but they often maintain \textit{diversity} in the population for a longer period of time. This allows generational GAs to easily escape local optima.

By contrast, \textit{steady-state} GAs replace only a few members of the population at a time. The parenthood selection mechanism typically selects a small number of highly-fit individuals to reproduce and generate a small number of children. In some versions, the children replace the weakest members of the population. Steady-state GAs often exhibit faster convergence rates than their generational counterparts since strong individuals quickly take over the population. However, it is more difficult for steady-state GAs to escape local optima.

Cellular GAs are a specialized version of generational GAs. In a cellular GA, the population is organized along a 2-dimensional grid. During the crossover phase, each individual selects a mate from the set of adjacent individuals. This mechanic simulates isolation by distance within the population. Discoveries of high-fitness individuals take time to propagate across the grid. This helps to prevent premature convergence and promotes niches of specialized subpopulations which cover a wider variety of optima \cite{15}.

The fitness landscape for QWOP is difficult to explore using most algorithmic optimization techniques. The solution space of QWOP gaits is enormous, and only a small fraction of the solution space results in gaits that do not immediately fail. Due to the game’s sensitive physics engine, small changes in control sequence can produce disastrous results. Genetic algorithms pose a promising technique for the automatic discovery of QWOP gaits given
their ability to explore large and jagged fitness landscapes.

\section{Related Work}

Much research has explored discovery and optimization of robotic gaits using genetic algorithms. Quadruped and hexapod robots are particularly common research platforms due to their physical stability and wide availability. Clune et al. \cite{5} created a genetic algorithm which evolves neural networks for quadruped gaits. Gallagher et al. \cite{8} used a similar approach to evolve gaits for a hexapod robot. Many other researchers have successfully employed genetic algorithms to evolve many-legged, neurocontrolled robots \cite{22} \cite{21} \cite{14}.

Rather than evolving weights for neural networks, many researchers tackle the problem more directly. Parker and Tarimo \cite{16} introduced the “Cyclic Genetic Algorithm” (CGA) and showed that it can effectively discover semi-optimal gaits for quadruped robots within 2000 generations. Their CGA encodes a solution as a sequence of control inputs to be sent to the robot, and the inputs are looped until evaluation has completed. Hornby et al. \cite{10} successfully employed genetic algorithms to automate the discovery of walking and running gaits on consumer-grade quadruped robots and included some of the resultant gaits in Sony’s AIBO robotic dog.

Genetic algorithms have also been applied to bipedal robots. The problem of evolving successful bipedal gaits can be particularly tricky compared to the quadruped or hexapod case. As Ray et al. \cite{19} note, “[W]alking with only two legs necessarily involves periods of time during which the robot is supported entirely by only one leg. Thus, bipedal gaits require more careful balance than gaits in which multiple legs are supporting the robot at any given moment.”

Wolff and Nordin \cite{24} ran an online genetic programming algorithm to optimize gaits for stability and straightness on a humanoid robot. Their algorithm refined gaits using on-board visual feedback. Zhang et al. \cite{26} discovered stable biped gaits for flat ground and for stair-climbing using a genetic algorithm to generate trajectories for the hip and ankle joints. Cheng et al. \cite{4} designed a GA capable of optimizing bipedal gaits for stability, speed, or inclined-surface walking. Many other researchers have successfully evolved human-like gaits using evolutionary computing \cite{25} \cite{6} \cite{1} \cite{3}.

The problem of evolving gaits for the simulated runner in QWOP is significantly easier than most of the gait evolution problems encountered in the literature. First, the QWOP runner interacts with a purely simulated 2-dimensional environment. QWOP gaits need not contend with unexpected obstacles, elevation shifts, or unexpected bumps in the walking surface. Second, the QWOP runner has only four degrees of freedom. The search space for QWOP gaits is much smaller than that of 3D bipedal or quadrupedal robots, which typically have 8 - 12 degrees of freedom.

At the time of this paper’s writing, only one publication exists which considers the specific problem of evolving gaits for QWOP. Ray et al. \cite{19} compare two GAs for QWOP. The first is a simple genetic algorithm (SGA) \cite{12} which represents solutions as sequences of key presses, holds, and releases. The second is a cellular genetic algorithm \cite{9} which represents solutions as sequences of characters from a 16-letter alphabet. Each letter in the alphabet corresponded to a certain configuration of keys to be held and released. In both cases, the authors used two-point crossover and uniform mutation. Fitness was assigned based on the speed that the simulated runner achieved. Experimental results indicated that the cellular genetic algorithm outperformed the simple genetic algorithm. The cellular genetic algorithm was able to evolve a stable knee-scooting gait similar to the strategy employed by many human players.

\section{Methods}

\subsection{QWOP Interface}

Individuals in our genetic populations are assigned fitness based on their performance in QWOP. Unfortunately, QWOP is a closed-source, browser-based game. It does not expose an API which could be used to programmatically play the game. Instead, we have designed
and implemented a Python program capable of automatically playing QWOP by controlling the mouse and keyboard. We call our software “Totter” \cite{13}.

To evaluate an individual’s fitness, Totter first uses the selenium \cite{11} library to open a web browser and navigate to Bennett Foddy’s website, where QWOP is hosted. The QWOP window is then resized and centered on the user’s primary monitor. Totter starts the game by programmatically moving the mouse to the center of the browser window and clicking to give it keyboard focus.

\begin{figure*}
    \centering
    \begin{subfigure}{0.48\textwidth}
        \includegraphics[width=\textwidth]{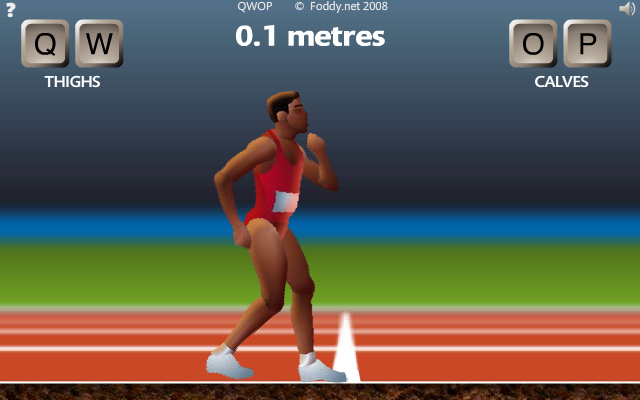}
    \caption{The simulated runner on the starting line}
    \label{fig1a}
    \end{subfigure}\hfill
    \begin{subfigure}{0.48\textwidth}
        \includegraphics[width=\textwidth]{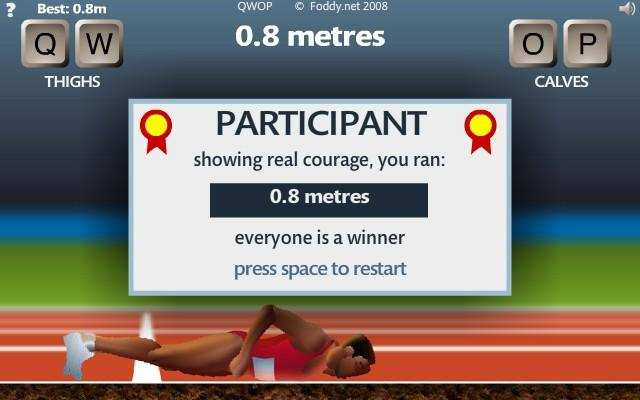}
    \caption{The “Game Over” screen after the runner falls}
    \label{fig1b}
    \end{subfigure}
    \caption{Screenshots from QWOP}
    \label{fig:1}
\end{figure*}

Totter then translates the individual’s genotype into a phenotype, which is a subroutine that executes a sequence of key presses, holds, and releases. The individual’s phenotype is executed in a loop using the pyautogui \cite{20} library. Execution terminates when one of the following conditions is met:

\begin{enumerate}
    \item The runner falls to the ground before the 100m mark and loses the game
    \item The runner reaches the 100m mark and wins the game
    \item The time limit expires
\end{enumerate}

Termination conditions 1 and 2 are detected using a color-sampling algorithm. QWOP’s “game over” screen, depicted in Figure \ref{fig1b}, contains a very different distribution of colors compared to the game’s normal background. Totter samples the color distribution of the window every 0.25 seconds, and terminates execution if the distribution matches the game over screen.

The third termination condition serves a dual purpose. First, it is possible to enter a sequence of inputs which causes the simulated runner to neither fall nor advance. If fitness evaluation were not time-limited, then such a sequence could stall the program indefinitely. Second, it is useful to limit the maximize time for a fitness evaluation. Evaluating an individual’s fitness is the bottleneck for performance in our algorithms. Fitness evaluations can take several minutes if a particularly slow but stable gait is discovered. Imposing a time limit allows evolution to continue at a reasonable pace.

After a termination condition is detected, Totter captures a small window of the screen which contains a description of the runner’s performance. Our program uses the Tesseract Optical Character Recognition \cite{23} library to convert the image into text data. Totter extracts the distance run and reports this value to the individual. The individual then updates its fitness. This process is repeated for each individual that a genetic algorithm needs to evaluate.

\subsection{Representation}

QWOP gaits can be described by a sequence of key presses, holds, and releases. We explored four different representations for QWOP gaits, each of which are described below. Regardless of representation, Totter forces key presses to be at least 50 milliseconds apart. This ensures that a computer-controlled player does not execute a key sequence that would be impossible for a human player. The experiment described in section \ref{sec:5.1} 5.1 is designed to assess the performance of each representation.

\subsubsection{Representation 1: Keystroke sequence}

The simplest possible representation encodes a gait as a sequence of keys to be pressed. Each individual is a variable-length string from the alphabet \{ Q, W, O, P\}. Each character in the sequence indicates that the corresponding key should be pressed for 150 milliseconds, then released. We refer to this representation as the “Keystroke” representation. An example individual is show in Figure \ref{fig:2}.

\begin{figure}[H]
    \centering
    \begin{tabular}{|c|}
    \hline
    {[}W, W, P, Q, P, W, O, O, P, Q, W{]} \\
    \hline
    \end{tabular}
    \caption{Example runner using Representation 1}
    \label{fig:2}
\end{figure}

The individual in Figure \ref{fig:2} would be translated as follows: Press W and hold for 150 ms, release W, press W and hold for 150 ms, release W, press P and hold for 150 ms, release P, press Q and hold for 150 ms, . . . and so on. This representation has two major benefits: it rarely contains sensitive patterns that could be disrupted by crossover and mutation, and it is highly interpretable. However, it is impossible to represent sequences with multiple buttons pressed at the same time, and it is impossible to vary the length of the button holds.

\begin{table*}
\centering
\caption{Mappings from alphabet to bitmask}
\label{tab:1}
\begin{tabular}{|c|p{-3em}|c|}
\cline{1-1} \cline{3-3}
\begin{tabular}{c|cccc}
Alphabet & Q & W & O & P\\
\hline
A & 1 & 1 & 1 & 1 \\
B & 1 & 1 & 1 & 0 \\
C & 1 & 1 & 0 & 1 \\
D & 1 & 1 & 0 & 0 \\
E & 1 & 0 & 1 & 1 \\
F & 1 & 0 & 1 & 0 \\
G & 1 & 0 & 0 & 1 \\
H & 1 & 0 & 0 & 0 \\
\end{tabular} & &
\begin{tabular}{c|cccc}
Alphabet &Q & W & O & P \\ 
\hline
I & 0 & 1 & 1 & 1 \\
J & 0 & 1 & 1 & 0 \\
K & 0 & 1 & 0 & 1 \\
L & 0 & 1 & 0 & 0 \\
M & 0 & 0 & 1 & 1 \\
N & 0 & 0 & 1 & 0 \\
O & 0 & 0 & 0 & 1 \\
P & 0 & 0 & 0 & 0 \\
\end{tabular} \\
\cline{1-1} \cline{3-3}
\end{tabular}
\end{table*}

\subsubsection{Representation 2: Keyup-Keydown Sequence}

The second representation expands on the search space by both allowing variable lengths of key presses and allowing multiple keys to be pressed at the same time. This representation encodes gaits as sequences of characters from the alphabet \{ Q, W, O, P, q, w, o, p, + \}. Under this representation, a capital letter means that the corresponding key should be pressed, a lowercase letter means that the corresponding key should be released, and a ‘+’ symbol indicates that the current configuration should be maintained for another 150 milliseconds. A 150 millisecond delay is inserted between interpretation of each character in the sequence. This representation is very similar to “Encoding 1” considered by Ray et al \cite{19}. We refer to it as the “Keyup-Keydown” representation. An example individual is shown in Figure \ref{fig:3}.

\begin{figure}[H]
    \centering
    \begin{adjustbox}{max width=0.48\textwidth}
    \begin{tabular}{|c|}
    \hline
    [P, p, +, W, Q, O, w, +, W, q, o, P, +, +, q, o, O, w, O, W, o] \\
    \hline
    \end{tabular}
    \end{adjustbox}
    \caption{Example runner using Representation 2}
    \label{fig:3}
\end{figure}

The individual in Figure \ref{fig:3} would be translated as follows: Press and hold P, release P, wait 150 ms, press and hold W, press and hold Q, press and hold O, release W, . . . and so on. Like representation 1, this representation is highly interpretable. Representation 2 allows individuals to produce more complicated patterns of keystrokes than representation 1, but it forces the GA to search a larger space of possible solutions. Another downside of this representation is that the control state at any given point on an individual using this encoding is highly dependent on its context within the sequence. Crossover and mutation operations are likely to produce “non-coding DNA” - patterns of inputs that cause no change in the current control state.

\subsubsection{Representation 3: Bitmask Sequence}

The third representation encodes a gait as a variable-length string of characters from the alphabet \{ A, B, C, . . . , P\}. Each character in the alphabet is mapped to a binary mask that sets the input state for all four keys. Table \ref{tab:1} shows the mapping of each letter in the alphabet to its corresponding bitmask.

Similarly to representations 1 and 2, each character in the sequence indicates that the corresponding control state should be held for 150 milliseconds. We refer to this representation as the “Bitmask” representation. An example individual is show in Figure \ref{fig:4}.

\begin{figure}[H]
    \centering
    \begin{tabular}{|c|}
    \hline
    [B, D, A, M, P, D, B, M, H, F, G, E, I, J, M, P]\\
    \hline
    \end{tabular}
    \caption{Example runner using Representation 3}
    \label{fig:4}
\end{figure}

The individual in Figure \ref{fig:4} would be translated as follows: press and hold Q, W, and O, then continue to hold Q and W while releasing O, then press and hold O and P, then release Q and W, then release O and P, . . . and so on. This representation allows individuals to press and hold multiple keys at the same time, but it does not allow for variable-length holds of a specific control sequence. Unlike representation 2, the Bitmask representation is not prone to heavy disruption by crossover and mutation operators, and will almost never produce non-coding DNA.

\subsubsection{Representation 4: Bitmask-Duration Sequence}

The final representation we consider is a simple modification of representation 3 which allows control states to be held for varying lengths of time. It encodes gaits as sequences of (\textit{X, duration}) pairs, where $X$ is one of the characters from the alphabet in table \ref{tab:1}, and duration is an integer in the range [$50,\infty$). An entry in the sequence indicates that control state $X$ should be held for \textit{duration} milliseconds. Figure \ref{fig:5} shows an example individual
encoded with this representation

\begin{figure}[H]
    \centering
    \begin{adjustbox}{max width=0.48\textwidth}
    \begin{tabular}{|c|}
    \hline
    [L, 117], [F, 433], [A, 350], [E, 440], [O, 468], [J, 349], [E, 76] \\
    \hline
    \end{tabular}
    \end{adjustbox}
    \caption{ Example runner using Representation 4}
    \label{fig:5}
\end{figure}

The individual in Figure \ref{fig:5} would be translated as follows: Press and hold W for 117 ms, release W, press and hold Q and O for 433 ms, press and hold W and P and hold for 350 ms, ... and so on. Encoding a gait as (\textit{bitmask, duration}) pairs allows the GA to attain the vast representational power of the “Keyup-Keydown” representation, without allowing for the potential of non-coding DNA. We refer to this representation as “Bitmask-Duration” representation.

\subsection{Fitness Function}

A naive approach to fitness assignment is to maximize distance achieved by the virtual sprinter. Such a fitness landscape selects for maximally stable individuals but ignores speed. This tends to produce runners who are very stable but also very slow. Another naive approach is to maximize running speed. But such an objective disproportionately allocates fitness to runners who take one lunging step at the beginning of the race and instantly crash.

To prevent over-rewarding falling lunges, other researchers \cite{19} assigned a fitness of zero to any runner who crashes. Our approach is slightly different. We use the following piece-wise function of distance ran and minutes elapsed:

\begin{equation*}
\begin{split}
F(distance,~&minutes) =  \\
& \begin{cases}
distance,                            & distance < 10\\
distance + \frac{distance}{minutes}, & distance \ge 10
\end{cases}
\end{split}
\end{equation*}

This function awards speed only for runners who are stable enough to advance past the 10-meter mark without falling. Applying the speed bonus in units of meters per minute ensures that speed and distance are on roughly the same scale. Unless otherwise noted, this is the fitness function used for all experiments in this paper.

\begin{figure*}
    \centering
    \begin{subfigure}{0.48\textwidth}
        \includegraphics[width=\textwidth]{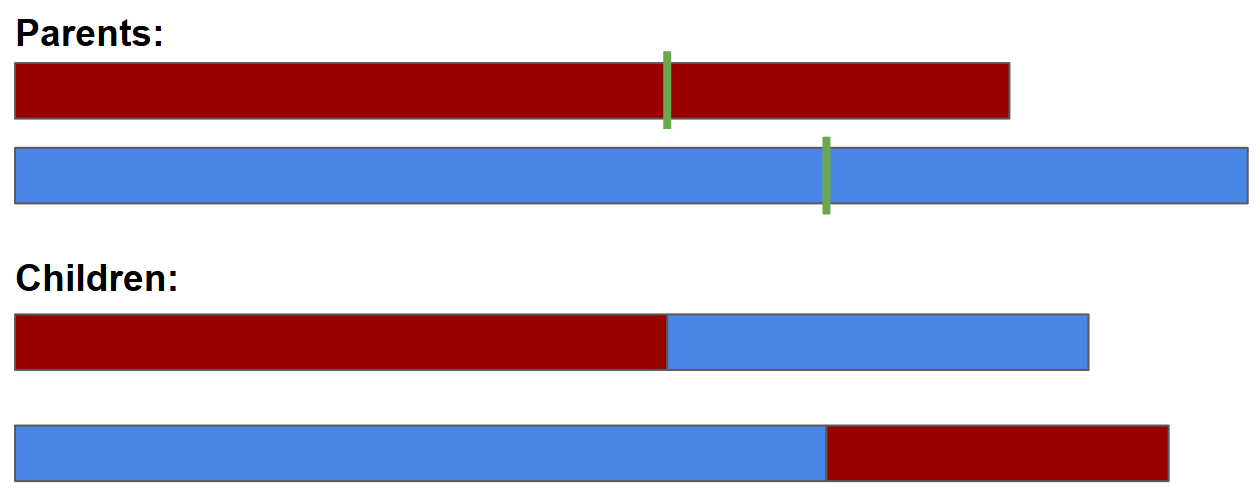}
    \caption{ Single-point “cut-and-splice” crossover}
    \end{subfigure}\hfill
    \begin{subfigure}{0.48\textwidth}
        \includegraphics[width=\textwidth]{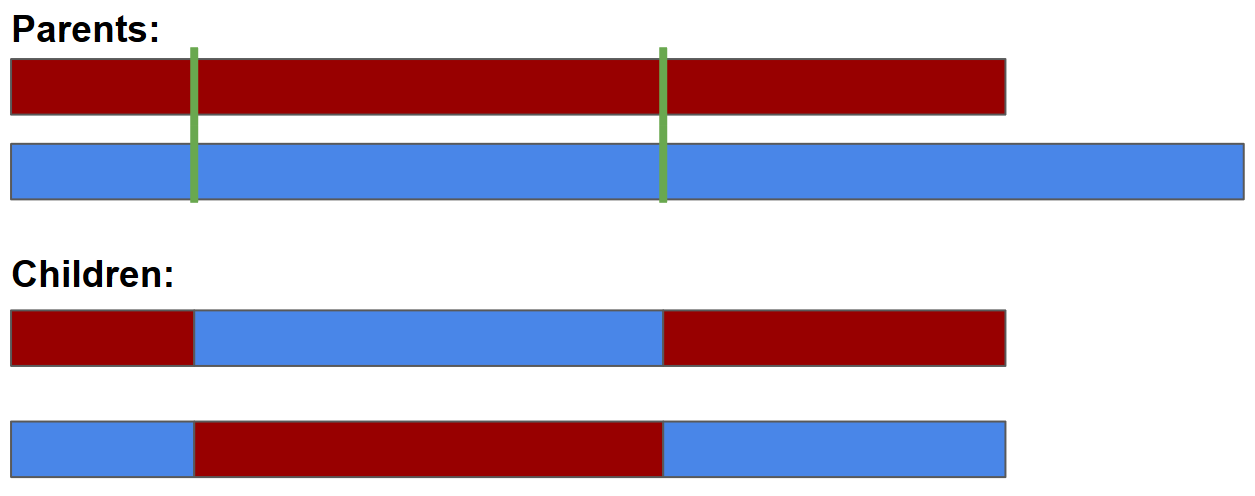}
    \caption{Two-Point crossover}
    \end{subfigure}
    \caption{Crossover operations}
    \label{fig:6}
\end{figure*}

\subsection{Initialization}
\label{sec:4.4}

We use two different initialization strategies in this paper. The first is random initialization. For representations 1, 2, and 3, an individual is assigned a random genome by first sampling an integer $k \in [20, 40]$, then choosing k letters from the representation’s alphabet. Representation 4 follows the above process, then randomly assigns a duration for each character by sampling
from the normal distribution $N(150, 25)$.

During experimentation, we noticed that nearly all randomly-generated QWOP gaits immediately fail. What’s more, an overwhelming majority of the children created by the crossover of unfit parents also immediately fail. Thus an initial population with no stable individuals sometimes fails to improve within a reasonable number of generations. Many GAs that used randomly generated initial populations were unable to evolve an individual that advanced more than one meter before falling.

To solve this problem, we introduced a second initialization strategy. We \textit{seed} the population by first generating a large pool of $M$ randomly-initialized individuals. Each individual in the pool is assigned a temporary fitness equal to the number of meters that individual is able to run within a 20-second time limit. The fittest N individuals are selected to be the initial population for the genetic algorithm. We refer to this strategy as “Seeding$_{M,N}$ initialization”.

\subsection{Selection and Replacement}

To select parents, we used tournament selection with tournaments with size equal to $\frac{1}{4}$ of the population size. When selecting individuals for replacement, we randomly chose one of the worst five members of the population. This greedy approach was chosen for its simplicity, elitism, and quick convergence.

\subsection{Crossover}

Except where otherwise noted, crossover was applied with a 90\% probability. In the cases where crossover was not performed, the two parents were carried forward as if they were children. If crossover was to be performed, we applied either single-point “cut-and-splice” crossover or two-point crossover with equal probability. Figure \ref{fig:6} shows a visual representation of these two operations.

\subsection{Mutation}

Except where otherwise noted, mutation was applied to each child’s genome with a 15\% probability. When mutation was to be performed, we applied one of the following operations, selected with equal probability:

\begin{enumerate}
\item Uniform replacement: change one of the characters in the sequence to another randomly selected character from the relevant alphabet
\item Insertion: randomly select a character from the relevant alphabet and insert it at a randomly-selected position in the genome
\item Swap: randomly select two characters in the sequence and swap their position
\item Deletion: remove a randomly-selected character from the sequence
\end{enumerate}

For representation 4, we also applied a Gaussian creep mutation to the \textit{duration} of a randomly selected (\textit{bitmask, duration}) pair from the sequence.

\section{Experiments}

The experiments we describe are designed to gauge the effectiveness of differing representations, initialization strategies, dynamic parameter control mechanisms, and various types of genetic algorithm. Because genetic algorithms are stochastic in nature, each experiment consists of several trials. Each trial starts with a freshly initialized population and runs until reaching 1000 fitness evaluations. In each trial, fitness evaluation for a each individual was subject to a time limit of 45 seconds. Unless otherwise noted, each trial uses the crossover and mutation operators described in Sections 4.6 and 4.7 with crossover probability fixed at 90\% and mutation probability fixed at 15\%.

The bottleneck in our algorithmic pipeline is fitness evaluation. The average time required to run a single trial for 1000 fitness evaluations was nearly six and a half hours. Furthermore, trials cannot be run in parallel because Totter requires control of the keyboard and mouse while running each trial. Because of the intense amount of time required for each trial, experiments were limited to five trials each. As such, the results we obtain should be viewed as preliminary and subject to more rigorous statistical significance testing in the future.

For each experiment, we report the mean best fitness (MBF), standard deviation in MBF ($\alpha_{MBF}$), and mean average fitness (MAF) achieved by each experimental configuration across all trials. Figures \ref{fig:7} - \ref{fig:10} plot MBF versus the number of fitness evaluations. The vertical lines in Figures \ref{fig:7} - \ref{fig:10} correspond to standard deviation in MBF across trials. The length of each vertical line equals the standard deviation in MBF measured at that particular generation.

\subsection{Representations}
\label{sec:5.1}

\begin{figure}
    \centering
    \includegraphics[width=0.48\textwidth]{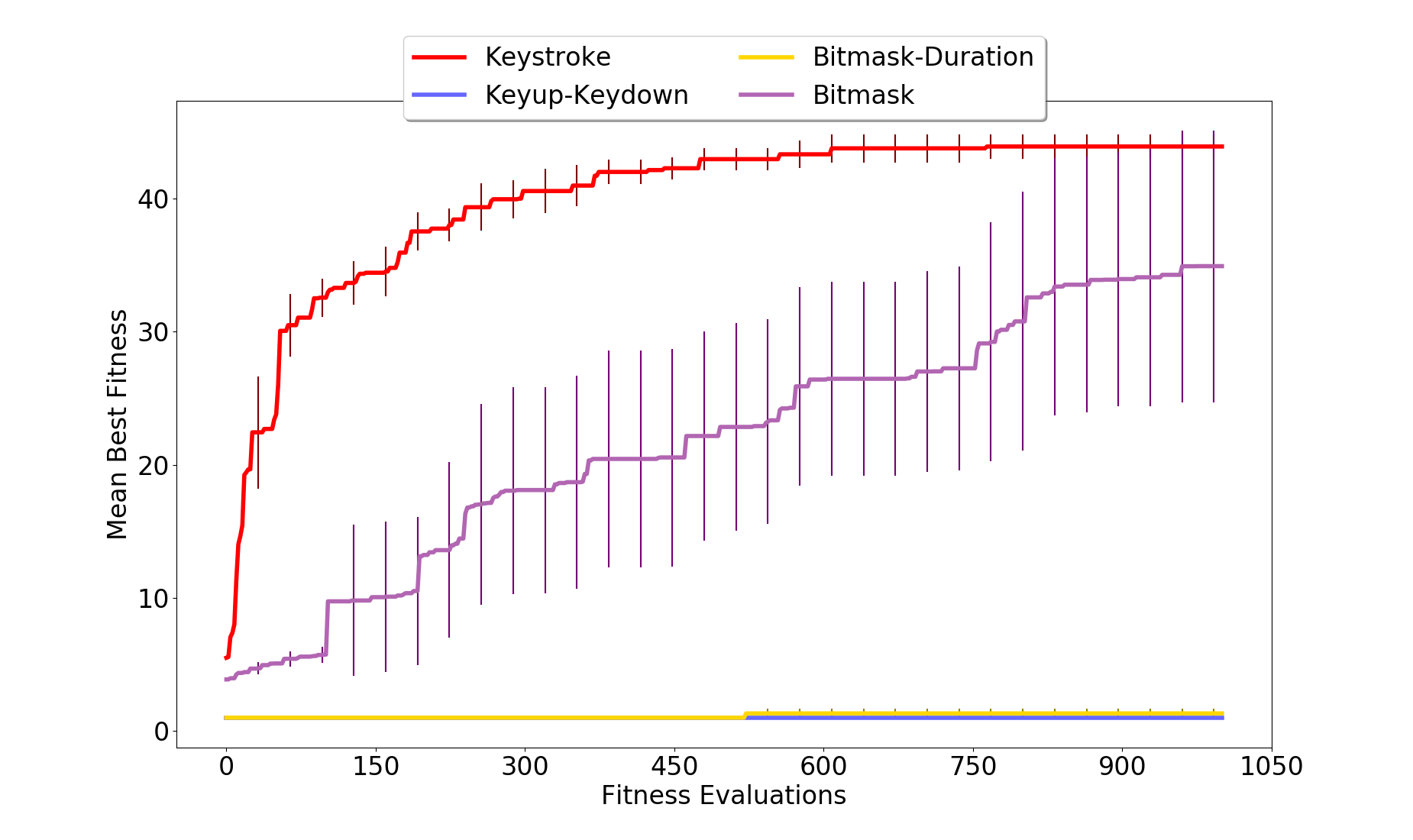}
    \caption{Fitness improvement for various representations}
    \label{fig:7}
\end{figure}

The first experiment tests the effectiveness of each representation. We vary representation while holding all other GA parameters constant. In each trial, we evolved populations of size 25 using a steady-state genetic algorithm. The population was initialized using Seeding$_{500,25}$ initialization as described in Section \ref{sec:4.4}. Figure \ref{fig:7} and Table \ref{tab:2} summarize the results of Experiment 1.

\begin{table}[htbp]
\centering
\caption{Results for Experiment 1}
\label{tab:2}
\begin{tabular}{r|ccc}
\textbf{Representation} & \textbf{MBF} & \textbf{MAF} & $\alpha_{MBF}$ \\
\hline
Keystroke        & 43.90 & 36.29 & 1.83  \\
Keyup-Keydown    & 1.0   & 0.93  & 0.0   \\
Bitmask          & 34.92 & 26.78 & 20.42 \\
Bitmask-Duration & 1.32  & 0.88  & 0.72  \\
\end{tabular}
\end{table}

The GAs using Keyup-Keydown representation and Bitmask-Duration representation fail to evolve solutions which are better than the best randomly-generated solutions. Both of these representations expand the search space to allow for variable-length holding of control states. We suspect that the GAs used for this experiment were unable to explore this expanded search space within 1000 fitness evaluations. These representations may do better when combined with more greedy operators or when initialized with a larger seeding pool.

By contrast, the GAs using Keystroke representation and Bitmask representation evolved populations that achieve respectable speed and stability. The Keystroke representation more reliably produced good runners, achieving both the highest MBF and MAF. From these results we can surmise that the ability to press multiple keys at once is relatively unimportant. The representation that achieved the best fitness is the one which most simplifies the search space.

The shape of each improvement curve is notable. The Keystroke GA improved quickly but “saturated” after 600 fitness evaluations. The Bitmask GA improved at a slower rate but did not saturate. It may be the case that the Bitmask GA would attain a higher fitness than the Keystroke GA if it were allowed to run for more than 1000 generations.

\subsection{Initial Population Seeding}

\begin{figure}
    \centering
    \includegraphics[width=0.48\textwidth]{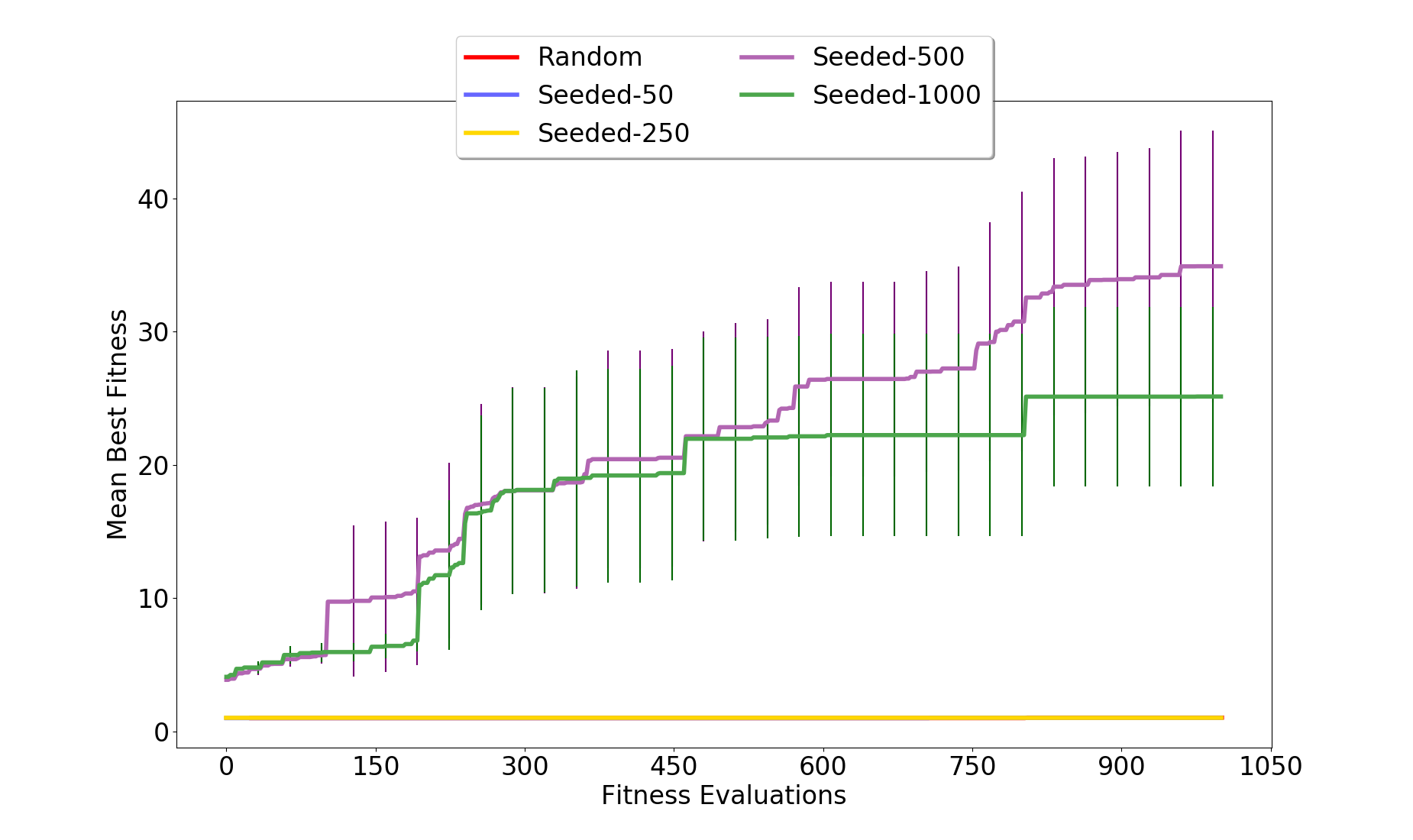}
    \caption{Fitness improvement with varying M}
    \label{fig:8}
\end{figure}

Our second experiment tests the effectiveness of the Seeding$_{M,N}$ initialization mechanism described in Section \ref{sec:4.4}. We compare the results of random initialization with the results obtained using Seeding$_{M,N}$ initialization with the following values of $M$: 50, 250, 500, 1000. This experiment ran steady-state genetic algorithms using Bitmask representation. Figure \ref{fig:8} and Table \ref{tab:3} summarize the results of Experiment 2.

\begin{table}[htbp]
\centering
\caption{Results for Experiment 2}
\label{tab:3}
\begin{tabular}{r|ccc}
\textbf{Initialization} \textbf{Strategy} & \textbf{MBF} & \textbf{MAF} & $\alpha_{MBF}$ \\ 
\hline
Random             & 1.04  & 0.88  & 0.09  \\   
Seeded$_{50,25  }$ & 1.02  & 0.90  & 0.04  \\   
Seeded$_{250,25 }$ & 1.0   & 0.94  & 0.01  \\    
Seeded$_{500,25 }$ & 34.92 & 26.78 & 20.42 \\ 
Seeded$_{1000,25}$ & 25.14 & 18.24 & 13.47 \\ 
\end{tabular}
\end{table}

These results demonstrate the importance of having better-than-random individuals in the initial population. When the population is randomly initialized, or when the population was seeded from a small pool of random individuals, the GA failed to improve upon the fitness of the initial population. When the GA was seeded from a pool of 500 individuals, it was able to make significant improvements to the best members of the initial population.

Increasing the size of the seeding pool past 500 seems to result in diminishing returns. The GA which drew from a seeding pool of size 1000 often got stuck in a local maximum and failed to achieve a fitness higher than the GA which used a smaller seeding pool. It may be the case that greedy selection in the initialization phase inhibits the GA’s ability to evolve innovative new solutions.

\begin{figure}[ht]
    \centering
    \includegraphics[width=0.48\textwidth]{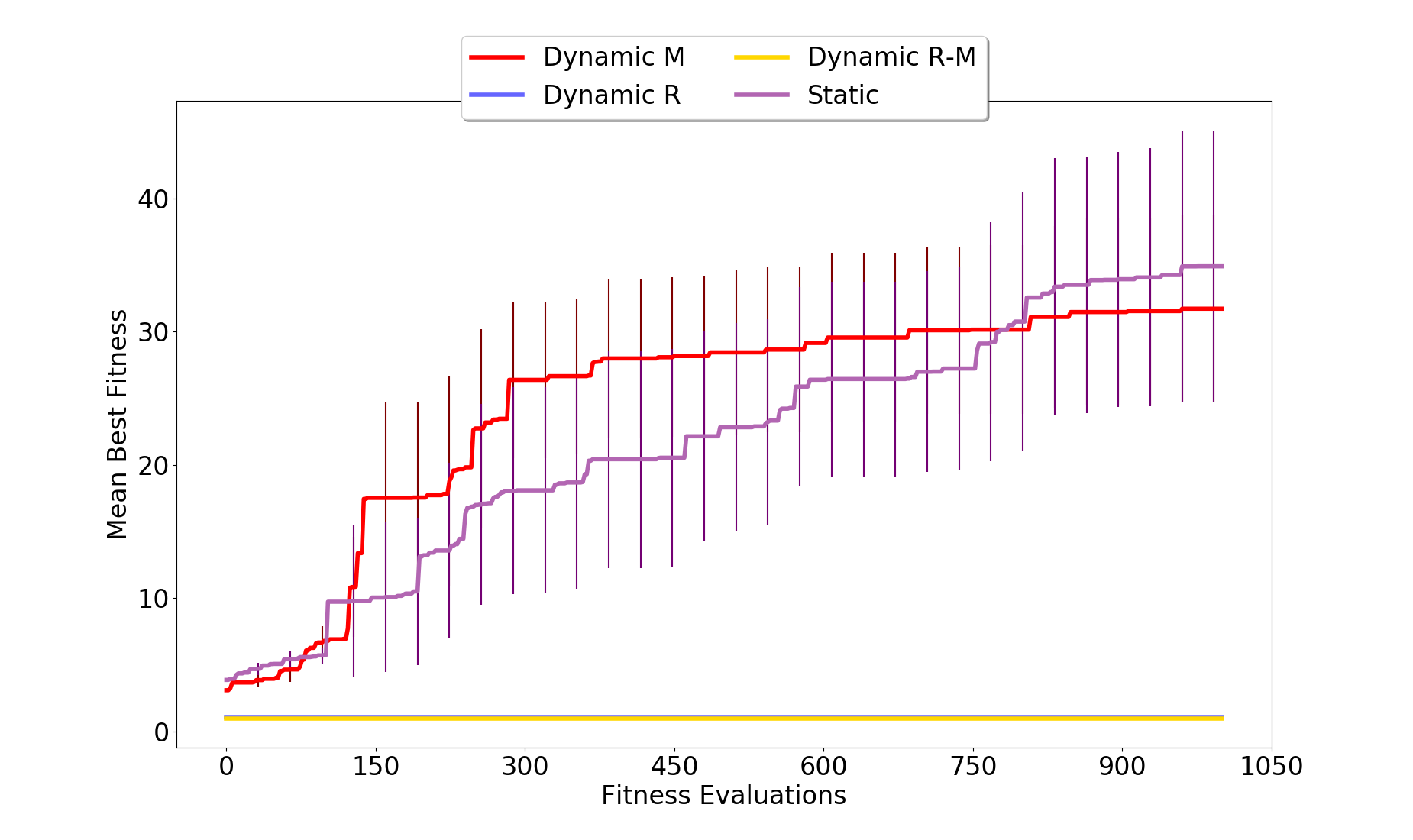}
    \caption{Fitness improvement for the GAs in experiment 3}
    \label{fig:9}
\end{figure}

\subsection{Dynamic Parameter Control}

In all experiments considered thus far, parameters such as the crossover rate, mutation rate, tournament size, and replacement operator have been kept constant throughout the run. Experiment 3 tests the effect of dynamically varying mutation rate and selection pressure with time.

First, we modify the steady-state GA by treating the mutation rate as a function of time. Generally speaking, a have a high probability of mutation during the early stages of evolution helps encourage exploration of the fitness landscape. In the late stages of evolution, a low mutation rate can help encourage convergence around the best solutions found so far. We use the following function to set the mutation rate: $m(t) = 1 - (0.85 * \frac{t}{T})$, where $t$ is the current generation and $T$ is the maximum number of generations.

Second, we modify selection pressure as a function of time. In previous experiments, we have randomly replaced one of the five worst members of the population. In experiment 3, we replace this mechanism with a probabilistic, time-dependent replacement strategy. Let $\bar{F}$ be the total fitness of the population. For an individual $x$ with fitness $f_x$, its probability of being selected for replacement is $P_r(x) = (1 - \frac{f_x}{\bar{F}})^{\alpha(t)}$, where $\alpha(t) = 0.5 + 1.5 \frac{t}{T}$. This replacement strategy is similar to inverse roulette-wheel selection, except that probabilities are scaled by a time-dependent factor. Early in the search, $\alpha(t)$ is less than one, causing selection pressure to be less steep. $\alpha(t)$ gradually increases as time advances, causing selection pressure to slowly rise.

We test these dynamic mechanisms by comparing among the following four genetic algorithms:

\begin{itemize}
\item Steady-state GA with static parameters (Static)
\item Steady-state GA with dynamic mutation only (Dynamic M)
\item Steady-state GA with dynamic replacement only (Dynamic R)
\item Steady-state GA using both dynamic replacement and dynamic mutation (Dynamic R-M)
\end{itemize}

In each case, we used a population of size 25, Bitmask representation, and Seeding$_{500,25}$ initialization. Figure \ref{fig:9} and Table \ref{tab:4} summarize the results of Experiment 3.

\begin{table}[htbp]
\centering
\caption{Results for Experiment 3}
\label{tab:4}
\begin{tabular}{r|ccc}
\textbf{Genetic Algorithm} & \textbf{MBF} & \textbf{MAF} & $\alpha_{MBF}$ \\
\hline
Static      & 34.92 & 26.78 & 20.42 \\ 
Dynamic M   & 31.73 & 25.50 & 14.04 \\ 
Dynamic R   & 1.4   & 0.51  & 0.01  \\ 
Dynamic R-M & 1.0   & 0.34  & 0.02  \\ 
\end{tabular}
\end{table}

Interestingly, the genetic algorithms that used the dynamic replacement mechanism both failed to evolve any better-than-random individuals. It may be that shallow selection pressure at the beginning of the search is harmful. Most initial populations have only one or two individuals who are better-than-random. If those individuals are replaced early, then the GA can fail to find any effective runners. This seems to be what has happened for the GAs who use dynamic replacement.

Dynamic mutation, on the other hand, does not seem to significantly help or hinder search progress. The GA only dynamic mutation improved more quickly than the GA with static parameters, but as the mutation rate declined, performance improvement significantly slowed and the static GA took over. Nevertheless, it seems that the dynamic mutation GA was able to more effectively explore the search space in early generations.

\subsection{Comparison of Genetic Algorithm Types}

Experiment 4 tests the effectiveness of the various types of genetic algorithm described in Section \ref{sec:2.2}. The results of Experiment 2 suggest that seeding the initial population helps kickstart evolution but may stall the GA in a local optimum. To control for this possibility, we run trials with each of the following configurations:

\begin{itemize}
\item Steady-state GA, random initialization (SS-R)
\item Steady-state GA, Seeding$_{500,25}$ initialization (SS-S)
\item Generational GA, random initialization (Gen-R)
\item Generational GA, Seeding$_{500,25}$ initialization (Gen-S)
\item Cellular GA, random initialization (Cell-R)
\item Cellular GA, Seeding$_{500,25}$ initialization (Cell-S)
\end{itemize}

\begin{figure}[ht]
    \centering
    \includegraphics[width=0.48\textwidth]{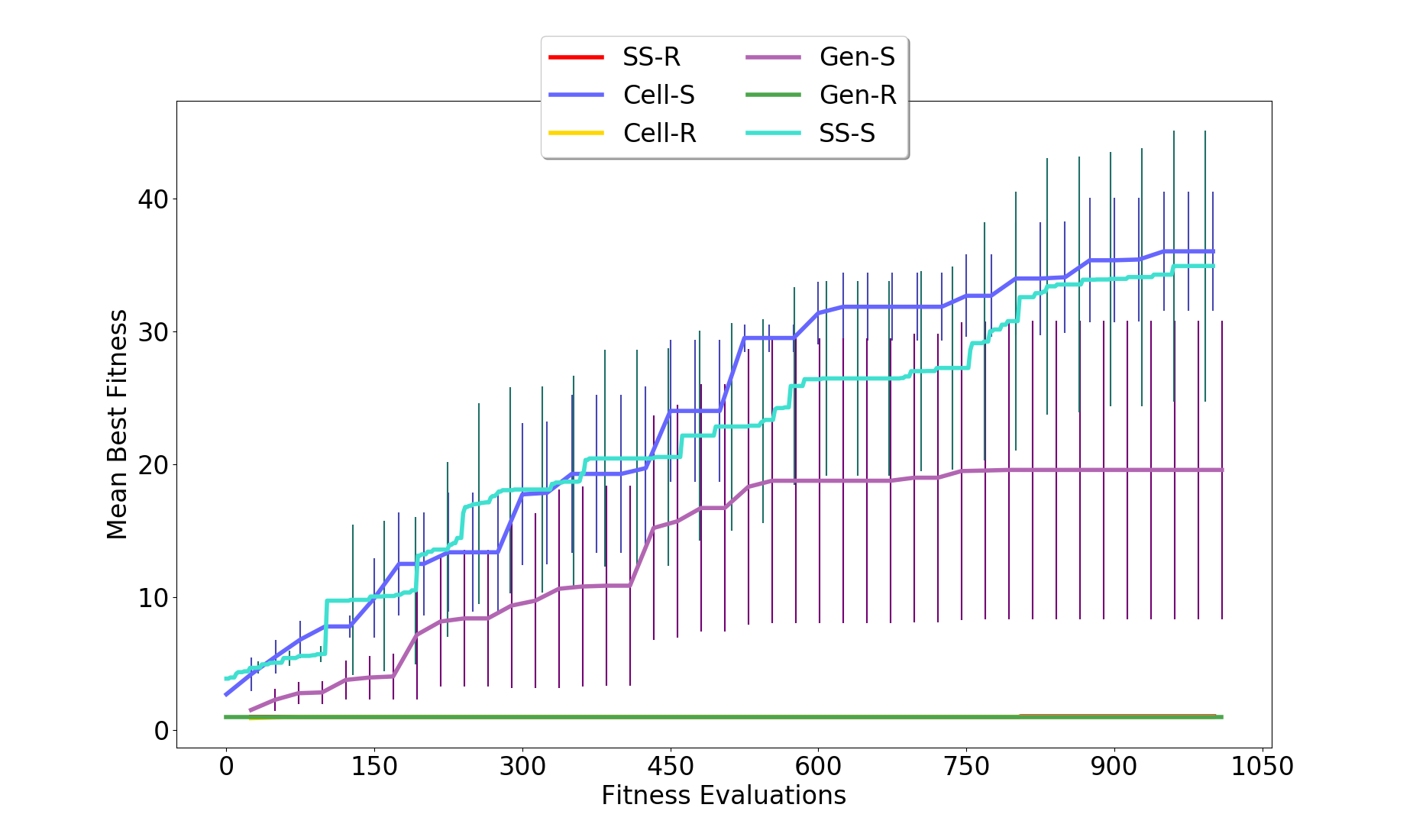}
    \caption{Fitness improvement for various types of GA}
    \label{fig:10}
\end{figure}

In each case, we used a population of size 25 and the Bitmask representation. Figure \ref{fig:10} and Table \ref{tab:5} summarize the results of Experiment 4.

\begin{table}[htbp]
\centering
\caption{Results for Experiment 4}
\label{tab:5}
\begin{tabular}{r|ccc}
\textbf{GA Strategy} & \textbf{MBF} & \textbf{MAF} & $\alpha_{MBF}$ \\
\hline
SS-R   & 1.04  & 0.88  & 0.09  \\
SS-S   & 34.92 & 26.78 & 20.42 \\
Gen-R  & 1.0   & 0.87  & 0.01  \\
Gen-S  & 19.57 & 14.67 & 22.43 \\
Cell-R & 1.0   & 0.44  & 0.02  \\
Cell-S & 36.02 & 7.48  & 8.95  \\
\end{tabular}
\end{table}

The cellular genetic algorithm most consistently yielded stable and fast gaits. Although MBF is similar between the steady-state and cellular algorithms, the cellular GA achieved a much lower standard deviation across trials. This phenomenon can likely be attributed to the increased diversity maintenance provided by the cellular paradigm. In the steady-state algorithm, a good individual quickly takes over the population. Thus, if the steady-state GA finds a local optimum early, it may converge to that optimum and never escape. By contrast, the discovery of a good individual takes many generations to propagate across the grid in the cellular GA. Notably, the MAF for the steady-state GA is quite high compared to the cellular GA. This indicates that our hypothesis is correct: by the final generation, the best individuals have taken over the steady-state algorithm, but the cellular algorithm continues to explore the search space.

These results also confirm the findings of Experiment 2. Seeding the population with better-than-random individuals is important regardless of the genetic algorithm used. When random initialization was used, no GA managed to evolve an individual which was better than-random.

\section{Conclusion}

This project experimented with several variants of genetic algorithm for the task of finding effective gaits for the game of QWOP. Our experiments suggest that steady-state and cellular genetic algorithms work well for this problem, particularly when quick convergence is desirable, but most algorithms need a population that already contains a few better-than-random individuals if evolution is to proceed successfully.

Many of the genetic algorithms we tested were able to evolve very stable gaits that, given enough time, would certainly complete the game by reaching the 100-meter mark. The best algorithms managed to find gaits that were both very stable and relatively fast. The gaits discovered by our algorithms are similar to the strategy that most novice- and intermediate-level human players employ. The gait begins by finding a stable configuration for the runner, usually by dropping to one knee or by spreading the runner’s legs. The runner maintains this position while kicking forward with his frong leg to generate forward momentum. A short video of the best runners we discovered is available for viewing using the following link: TODO: youtube video. The Python source code used to evolve runners and generate the figures in this paper is available at the following link: \href{https://github.com/zachdj/totter}{https://github.com/zachdj/totter}.

Although our algorithm finds relatively good solutions, it does not find gaits which approach the aptitude of the best human players. Humans who compete for the world record employ a realistic sprinting gait that requires careful balance, impeccable timing, and improvisation. Human-like control of the runner is unlikely to be achieved by a genetic algorithm similar to any considered in this paper. A system that can respond dynamically to sensory feedback would have a better chance of success than a simple sequence of inputs that is continuously looped.

Our work could be expanded in several interesting ways. First, the experiments in this paper only scratch the surface of the multitude of genetic algorithm variants which could be brought to bear on this problem. Our experiments also considered only one type of crossover and mutation operator. It may be the case that domain-specific operators would exhibit better performance. As noted by Ray et al. \cite{19}, the GA would almost certainly benefit from a fitness function that provided more information about the current state of the runner, such the configuration of his legs, arms, and body relative to the ground at any point in time.

As previously noted, the immense time required for fitness evaluation limits both the number of experiments that can be considered and the number of generations which a GA can be allowed to run. To find near-optimal gaits, a GA would likely need months of real-time execution. The temporal costs of evolution could be dramatically lowered by employing a surrogate fitness function. It may be possible to build a simulation of QWOP that mimics the real game closely but gives the programmer more fine-tuned control over the in-game clock. This would allow several thousand times more fitness evaluations per second.

Considering QWOP’s unpredictable ragdoll physics engine and deviously challenging control scheme, the gaits achieved by the algorithms in this paper are satisfactory. Our results can be considered a proof-of-concept in applying evolutionary computing to challenging games that require careful timing of controls.

\nocite{*}
\bibliographystyle{IEEEtran}
\bibliography{references}

\end{document}